# Online Signature Verification using Deep Representation: A new Descriptor

Mohammad Hajizadeh Saffar[1]. Mohsen Fayyaz[2]. Mohammad Sabokrou[3]. Mahmood Fathy[1]

[1]School of Computer Engineering, Iran University of Science & Technology, Tehran, Iran

[2]Faculty of Computer Science, University of Bonn, Bonn, Germany

[3]Computer Science department, Fundamental Sciences (IPM), Tehran, Iran

**This paper presents an accurate method for verifying online signatures. The main difficulty of signature verification come from: (1) Lacking enough training samples (2) The methods must be spatial change invariant. To deal with these difficulties and modeling the signatures efficiently, we propose a method that a one-class classifier per each user is built on discriminative features. First, we pre-train a sparse auto-encoder using a large number of unlabeled signatures, then we applied the discriminative features, which are learned by auto-encoder to represent the training and testing signatures as a self-thought learning method (i.e. we have introduced a signature descriptor). Finally, user's signatures are modeled and classified using a one-class classifier. The proposed method is independent on signature datasets thanks to self-taught learning. The experimental results indicate significant error reduction and accuracy enhancement in comparison with state-of-the-art methods on SVC2004 and SUSIG datasets.**

## I. INTRODUCTION

Authentication has been known as an intrinsic part of social life. Recent years have seen a growing interest toward personal identity authentication. Increasing security requirements have placed biometrics at the center of so much attention. Biometric technology has become an important field in verifying people and has been used in people identification and authentication. The term biometric refers to individual recognition based on a person's distinguishing characteristics [14].

Recognition refers to identification and verification tasks. Identification specifies which user provides a given biometric parameter among a set of known users. Therefore, the input used for identification only contains genuine data. However, verification determines if the given biometric parameter is provided by a specific known user or is a forgery.

Handwritten signature recognition is one of the most common techniques to recognize the identity of a person. However, when dealing with signatures, most of the proposed systems focus on verification rather than identification because of daily usage of signature verification systems [6].

There are two types of signature verification: Offline (static) and Online (dynamic) verification. In the offline setting, the shape of the signature has been captured. Therefore, in offline verification systems, input data contains $x$, $y$ coordinates of signatures. However, in the online setting, the system uses devices for capturing additional information while the user is signing [26]. Online signatures have extra information for extraction such as; time, pressure, pen up and down, azimuth, etc.

Generally, the Verification approaches which are used in previous researches can be described in three categories [14]:

1) *Template Matching*: A questioned sample has been matched against templates of signatures, such as Euclidean distance and Dynamic Time Wrapping (DTW) [2, 6, 26].
2) *Statistical*: In this approaches, distance-based classifiers can be considered, such as Neural Networks [17] and Hidden Markov Models (HMM) [6, 9].
3) *Structural*: This approach is related to structural representations of signatures and compared through graph or tree matching techniques [3].

Recently, *deep learning* provides state-of-the-art results for various biometric systems such as; iris [20], face and fingerprint [22] and finger-vein [7].

In this paper, a signature verification system based on deep learning has been proposed. A sparse linear auto-encoder has been implemented to learn the signature pattern of each user by learning features based on an *unsupervised self-taught* method. This feature learning is done on a large number of unlabeled signatures which are provided in ATVS dataset. As the number of labelled signature samples are limited, so learning the features on labelled ones is not feasible and the self-taught is a good choice for dealing with this restriction. Furthermore, one-class classifier has been used for classifying test signatures. The results of this paper confirm that the learned features are more discriminative rather than state-of-the-art methods where handcrafted features have been used. As the best of our knowledge, we are the first in introducing a descriptor for verifying the signatures using deep learning. The main contributions of this paper are three folds:

1) We introduce an efficient descriptor for online signatures, which can be applied in different datasets. Our experiment confirms our claim that we have achieved state-of-the-art results in two classic benchmarks.
2) We consider an online signature as an image with two channels (one channels is related to time information and another is related to pressure.)
3) We propose a one-class classifier to reject (i.e. detect) the forgery signature as an outlier.

This paper is organized as follows: Section II, presents previous work have been done in the field of signature verification. Section III introduces the adopted methodology for system architecture while section IV presents the proposed system with details. Experimental results and their comparisons have been described in section V. Finally, section VI presents the conclusion for this paper and suggestions for future work.

## II. RELATED WORK

Most recent approaches in the field of online signature verification have been described in [14, 15, 33]. The process of signature verification is usually divided into three phases:

### 1. Pre-processing

The signature dataset must take some pre-processes since there is no guarantee that different signatures of one user will always be the same. Several processes have been proposed for this phase, which generally consist of smoothing, rotation and normalization.

Cubic splines can be employed for smoothing purpose to solve the jaggedness in the signatures. Signatures can become rotation-invariant by rotating each one based on orthogonal regression (Eq. 1) [26].

$$\bar{\theta} = tg^{-1}\left(\frac{s_y^2 - s_x^2 + \sqrt{(s_y^2 - s_x^2)^2 + 4*(cov_{(x,y)})^2}}{2*cov(x,y)}\right) \quad (1)$$

Where $s_x$ and $s_y$ are variance and $cov_{x,y}$ is covariance of the horizontal and vertical components.

The signatures of one person must have the same size for better performance. The horizontal and vertical components of signatures can be normalized to make a standard size (Eq. 2, 3) [2].

$$x_n = \frac{x - \min(x)}{\max(x) - \min(x)} * 100 \quad (2)$$

$$y_n = \frac{y - \min(y)}{\max(y) - \min(y)} * 100 \quad (3)$$

Where $x$ and $y$ are original and $x_n$ and $y_n$ denote the normalized coordinates.

### 2. Feature Extraction

Feature selection and feature extraction play an important role in verification systems. Many studies have been performed in the field of feature selection to choose the best set of features for extraction. List of common features have been described in Table 1 [26].

Table 1. List of common features

| # | Description |
|---|---|
| 1 | Coordinate $x(t)$ |
| 2 | Coordinate $y(t)$ |
| 3 | Pressure $p(t)$ |
| 4 | Time stamp |
| 5 | Absolute position, $r(t)=square\_root(x^2(t),y^2(t))$ |
| 6 | Velocity in x, $v_x(t)$ |
| 7 | Velocity in y, $v_y(t)$ |
| 8 | Absolute velocity, $v(t)=square\_root(v_x^2(t),v_y^2(t))$ |
| 9 | Velocity of r(t), $v_r(t)$ |
| 10 | Acceleration in x, $a_x(t)$ |
| 11 | Acceleration in y, $a_y(t)$ |
| 12 | Absolute acceleration, $a(t)=square\_root(a_x^2(t),a_y^2(t))$ |

Furthermore, some non-common features have been described in other papers [1-3, 23, 24, 27, 31]. Recently, instead of handcrafted features in traditional biometric authentication systems for face, iris and fingerprint, some more discriminative features which are provided using deep learning are exploited [22, 30].

### 3. Classification

After the feature extraction phase, the system must create a model from reference signatures. For classification phase, each signature must be compared against reference signatures and by calculating the distances between test and reference signatures, the system can decide to accept or reject the test signature.

As mentioned, in daily usage of authenticating systems such as banking systems, handwritten signature of users has been used to verify the identity of official documents. In these sets of problem, the main goal is verifying whether a signature belongs to one identified person or not. In contrary to multi-class classifiers, the aim for one-class classifiers is distinguishing one type of class (target) from other classes (outlier). Thus, for classifying a signature as genuine or forgery, one-class classifiers have been commonly used [29] to divide the set into two categories: target and outlier (Figure 1). As it shown, detecting the random forgery ones is very easier than skilled forgery ones.

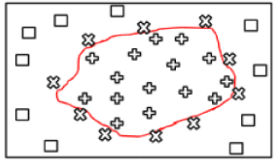

Figure 1 Example of signature model for each user. The red line is the boundary of genuine samples with skilled and random forgery signatures

Jain and Gangrade [17] proposed a system by using angle, energy and chain code features to differentiate the signatures. In this approach, a Neural Network has been applied for classification.

Faundez-Zanuy [6] studied four pattern recognition algorithms for online signature recognition: Vector Quantization (VQ), Nearest Neighbor, Dynamic Time Warping (DTW) and Hidden Markov Model (HMM). The author proposed two methods based on VQ and Nearest Neighbor.

Rashidi, et al. [26] evaluated 19 dynamic features viewpoint classification errors and discrimination capability between genuine and forgery signatures. They used a modified distance of DTW for improving performance of verification phase.

Ansari, et al. [2] presented an online signature verification system based on fuzzy modelling. The point of geometric extrema has been chosen for signature segmentation and a minimum distance alignment between samples has been made by DTW techniques. Dynamic features have been converted to a fuzzy model and a user-dependent threshold used for classification.

Barkoula, et al. [3] studied the signatures Turning Angle Sequence (TAS), the Turning Angle Scale Space (TASS) representations, and their application to online signature verification. In the matching stage, the authors have employed a variation of the longest common sub-sequence matching technique.

Yahyatabar, et al. [31] proposed a method based on efficient features defined in Persian signatures. A combination of shape based and dynamic extracted features has been applied and a SVM has been used for classification phase.

Alhaddad, et al. [1] explored a new technique by combining back-propagation Neural Network (BPNN) and the probabilistic model. BPNN has been used for local features classification, while probabilistic model has been used to classify global features.

Mohammadi and Faez [23] proposed a method based on the correspondence between important points in the direction of wrap for the time signal provided to maximize the distinction between the genuine and forged signatures.

Napa and Memon [24] Presented a simple and effective method for signature verification in which an online signature is represented with a discriminative feature vector derived from attributes of several histograms that can be computed in linear time. For testing phase, the authors proposed a method on finger drawn signatures on touch devices by collecting a dataset from an uncontrolled environment and over multiple sessions.

Souza, et al. [29] proposed an off-line signature verification system, which uses a combination of five distance measurements, such as; furthest, nearest, template and central using four operations: product, mean, maximum, and minimum as a feature vector.

Fallah, et al. [5] presented a new signature verification system based on Mellin transform. The features have been extracted by Mel Frequency Cepstral Coefficient (MFCC). Neural Network with multi-layer perception architecture and linear classifier in conjunction with Principal Component Analysis (PCA) have used for classification.

Iranmanesh, et al. [16] proposed a verification system by using multi-layer perceptron (MLP) on a subset of PCA features. This approach used a feature selection method on the information that has been discarded by PCA, which significantly reduced the error rate.

Cpałka, et al. [4] explored a new method by using area partitioning of high and low speed of the signature and high and low pen's pressure. The template for each partition has been generated and by calculating the distance between signatures and template in each partition, a fuzzy classification has been implemented to classify the signatures.

Lopez-Garcia, et al. [21] presented a signature verification system implemented on an embedded system. In this approach, a template for each user has been generated and a DTW algorithm has been used for distance calculation. Finally, the features extracted and passed through a Gaussian Mixture Model (GMM) to calculate the similarity between the test signature and the generated template.

Gruber, et al. [12] proposed a technique based on Longest Common Sub-sequences (LCSS) detection. Authors have used a LCSS kernel of SVM for classifying the similarity of signature time series.

### III. METHODOLOGY

Deep learning (*Feature Learning* or *Representation Learning*) is a new era of machine learning which aims to learn the high-level features from raw data to achieve a better performance in classification tasks. Deep learning is part of a field of machine learning methods based on learning representation of data [28].

Feature learning tries to learn discriminative features autonomously which is considered as one of its advantages. The other advantage of feature learning process is its capability to be completely unsupervised.

One of the scopes of machine learning, which plays a key role in deep learning, is *self-taught learning*. The main promise of self-taught learning is using unlabeled data in supervised classification tasks [25]. The key point of such algorithms is that unlabeled data are not supposed to follow the same class labels. Indeed, unlabeled data are exploited to teach the system recognizing patterns or relations for the supervised learning task. In summary, self-taught learning learns a concise, higher-level feature representation of the raw data using unlabeled data. This brings an easier classification task by having features that are more significant [25].

In following of this section, we explain the auto-encoder architecture, which is used for learning and extracting sparse and discriminative features of signatures. Furthermore, we explain how convolution and pooling techniques are exploited to the extracted features to become spatial-changing invariant.

1. Auto-encoder

Auto-encoder is one of the unsupervised feature learning tools. There is one kind of auto-encoder algorithms, which is based on multi-layer perceptron neural networks. In contrary to traditional neural networks, MLP based auto-encoders are unsupervised learning algorithms which try learning weights of each layer to set the output values to be equal to the inputs. The structure of auto-encoder is shown in Figure 2.

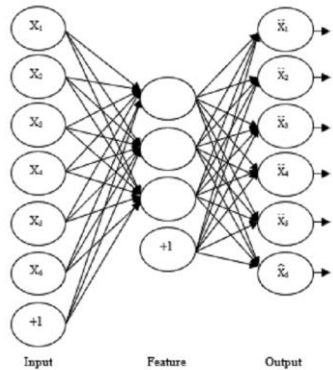

Figure 2 Architecture of an auto-encoder for learning kernel of convolutional layer

Suppose *x* is the set of input features. To learn features from input features, the basic auto-encoder with regularization term to prevent over-fitting, attempts reconstructing input features by minimizing following cost function (Eq. 4):

$$J(W,b) = \arg\min_{W,b} \frac{1}{m} \sum_i \|h_{w,b}(x^{(i)}) - x^{(i)}\|^2 + \lambda \sum_l \sum_{i,j} (W_{i,j}^l)^2 \quad (4)$$

Where *x* is the input features for a training example, *w* is the weight matrix mapping nodes of each layer to next layer nodes, and *b* is a bias vector.

The cost function of auto-encoder mentioned in (Eq. 4) only focuses on the differences between input and output data of auto-encoder. This brings a network with the ability of representing raw data with learned feature without any guarantee of having sparse represented features, which plays a key role in classification tasks. In order to learn features that are more effective and having a sparser set of represented features, the sparsity constraint can impose on the auto-encoder network. The objective function is as follows (Eq. 5-7):

$$J_{Sparse}(W,b) = J(W,b) + \beta \sum_i KL(\rho \| \hat{\rho}_j) \quad (5)$$

$$KL(\rho \| \hat{\rho}_j) = \rho \log \frac{\rho}{\hat{\rho}_j} + (1-\rho) \log \frac{1-\rho}{1-\hat{\rho}_j} \quad (6)$$

$$\hat{\rho}_j = \frac{1}{m} \sum_i [a_j^2(x^{(i)})] \quad (7)$$

Where $KL(p \| p_j)$ is the Kullback-Leibler (KL) divergence between a Bernoulli random variable with mean *p* and a Bernoulli random variable with mean $p_j$, which is the average activation of hidden unit *j*. In addition, *beta* is the weight of the sparsity penalty term, lambda is the Weight decay parameter and *p* is the Sparsity parameter, which specifies the desired level of sparsity.

A *sparse auto-encoder* model can effectively realize feature extraction and dimension reduction of the input data, which play a vital role in classification tasks [30].

2. Convolution and Pooling

Raw input data are usually stationary. In other words, randomly selected parts of the data have the same statistics. This characteristic shows that not all the features are useful. It is obvious that having more features results in increasing the computational complexity especially in a classification task. In order to avoid high complexity, redundant data have been neglected by picking up random patches of raw data and convolving them. After obtaining convolved features, pooling method can be exploited in order to obtain pooled convolved features (Figure 3).

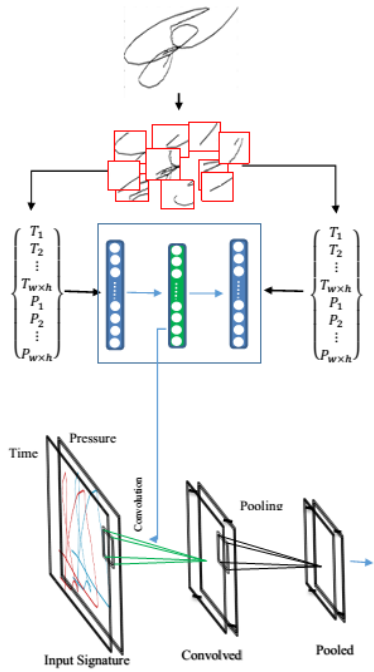

Figure 3 *Top*: Learning features from ATVS dataset. *Bottom*: Convolving and pooling the learned feature for representing the signatures

## IV. PROPOSED SYSTEM

One of the important problems in signature verification is choosing features due to diverse difficulties in signature verification, such as; differences between same user signatures, different circumstances of signing, various shapes of signatures, etc. Among these, exploiting an unsupervised feature learning method results in system compatibility improvement with various types of signatures and automatic feature selecting from signatures. To achieve more discriminative features, each signature has been considered as an image with two channels where the intensity of these channels are the pressure and time of each position of signature. First, an efficient descriptor has been learned for describing the signatures. This procedure is done based on self-thought learning using ATVS dataset. Then, for each user the training samples have been described using the learned features and a reference model has been fitted on the training user's signatures. In test phase and for verifying a signature, it has been checked with all models. If it has not be fitted to any of defense models of users, the signature is labelled as forgery, otherwise it is labelled as genuine for a specific user. Figure 4 shows an overview of proposed approach for verifying the user's signatures.

The proposed signature verification system comprises two main steps:

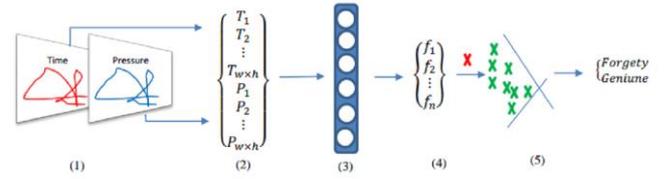

Figure 4 Workflow of signature verification. Left to right: (1) A test signature which is represented as an image with two channels: pressure and time. (2) Reshape the input sample to a vector. (3) The vector is represented by an auto-encoder where is pre-trained on ATVS dataset. (4) The input signature is described by n features. It is the output of auto-encoder (5) Classify the sample using one-class classifier that is learned by training samples of each user and make the final decision

### Step 1: Learning a signature descriptor

First step consists of creating signature descriptor based on self-tough learning. All signature samples in ATVS dataset have divided to 17500*64 Patches with the size of 10*10 and given to a sparse auto-encoder. After the auto-encoder is completely learned, it can be used as an efficient feature extractor from the signatures patches.

### Step 2: Creating references models for users

In the second step, all signatures have divided into small patches and they have described by the learned descriptor. The descriptions of small patches have pooled and the mean of them (i. e. mean of the features which are extracted from each sample) have considered as the descriptions of the signatures. Based on the explained procedure, all training signatures for each user have been described and a reference signature model for each user has been created. These models are considered as a set of one-class classifiers.

As described, in the first step, features are learned by an auto-encoder. In this step, an unlabeled dataset, which is discretized from train and test datasets, is used based on self-taught method. In the next step, a reference model of the system is built using classified represented data from user's reference signatures. These two steps are parts of the system training phase [24]. Finally, in verification phase, which is system-working section, new unknown signatures are compared against the system reference model (classified data) to be verified. There are three principal parts among described steps, which are pre-processing, feature learning using auto-encoder, and classification. These parts are explained as follows:

1. Pre-processing

As mentioned, in the pre-processing phase, normalizing size of the signature is the first step. This aim can be achieved by scaling the signature size. At the next step, the mean of the data must become equal to zero for data normalization.

Signatures data in datasets are based on time, pressure, pen up/down, etc. in *x, y* positions. To make representation become similar to reality, points of signatures have been continued. This object achieved by using time of the points to observe the sequence of data and pen up/down to check if the pen has gone up, the point must be separated from the next one. Finally, signatures have been represented base on two layer: pressure and time.

*Principal Component Analysis* (PCA) is an algorithm that reduces dimensions of signature data and can be used to significantly speed up unsupervised feature learning algorithm. Since the system is trained based on signature images, adjacent pixel values are highly correlated. Whitening can make the input less redundant, the features become less correlated with each other and all become the same variance. Therefore, these two algorithms have been used to reduce the dimensions.

2. Feature Learning using Auto-encoder

For learning features from signatures, a linear auto-encoder with sparsity have been used. The signature has been set for input and output and auto-encoder has been checked for mapping input to output. This auto-encoder has been designed based on gradient descent.

Unsupervised learning algorithms have high computational cost. In order to increase performance of learning phase, raw data (large patch of a signature) has been divided into small patches and have been used in feature learning phase as input. Then, learned features have been convolved with large patch. After obtaining features using convolution, mean pooling method has been exploited in order to obtain pooled convolved features. These pooled features have been used for classification.

3. Model creation and classification

The significant issues of classification in this type of problems are differences between same user's signatures, diverse circumstances of signing, low amount of signature samples, and forgery signatures. For resolving such issues, selecting an appropriate classifier is very important.

In the proposed system, the one-class classifier has a target class, which is the class of the user whose signature is being compared with input signature, and the outlier class is other user's sample signatures. As a result, the classifier must create a model of target class for each user.

## V. EXPERIMENTAL RESULTS

In the evaluation process of proposed approach, test signatures have been comprised by comparing their features against reference signatures. In this section, short description of benchmarks and evaluation parameters have been described. In addition, two main steps of the experiments have been explained.

1. Benchmarks

For evaluation of the proposed approach, three public datasets have been used which are *SVC2004* [32], *SUSIG* [19] and *ATVS* [10, 11]. The structure of the mentioned datasets have been explained as follows:

*A. SVC2004*

This is the first international signature verification competition. SVC2004 main dataset has 100 sets of signature data. SVC2004 public dataset, which has been released before the competition, consists of 40 signature sets. Each set includes 20 genuine signatures from each contributor and 20 skilled forgeries from at least four other contributors (Figure 5).

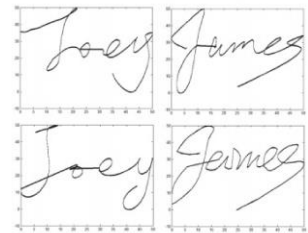

Figure 5 Examples of Genuine (first row) and Forgery (second row) signatures in SVC2004 dataset

In data collection process of signature sets, contributors were asked to contribute 20 genuine signatures in two sessions during two weeks. At least four other contributors forged the skilled forgeries for each contributor's signature.

In SVS2004 dataset, each signature includes a sequence of points, which contains *x, y* coordinates, time and pen up/down, azimuth, altitude and pressure.

*B. SUSIG*

Sabanci University Signature database (SUSIG) is a set of online signatures, which aim is overcoming some of the shortcomings of its contemporary datasets.

The SUSIG dataset consists of two sub-corpora, which are visual and blind (Figure 6).

In blind sub-corpus data collection, the process has been done on a tablet without visual feedback. It consists of 100 contributors. First group of 30 contributors provided eight genuine signatures, while the other 70 contributors provided 10 genuine signatures each.

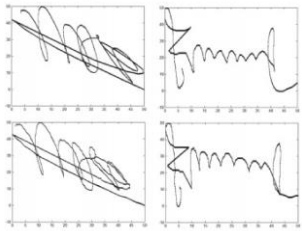

Figure 6 Examples of Genuine (first row) and Forgery (second row) signatures in SUSIG dataset

In visual sub-corpus data collection, the process has been done on a tablet with a LCD, which provided visual feedback to the contributors while they were signing signatures. Visual sub-corpus data were collected in two separate sessions. Each contributor has provided 20 samples of his/her signature.

*B. ATVS*

All two mentioned (SVC2004 and SUSIG) are human made datasets. In contrast, synthetic signature datasets have good approaches for simulation of real signatures, which involves the effect of real situation of sampling. ATVS dataset is one of the synthetic datasets (Figure 7).

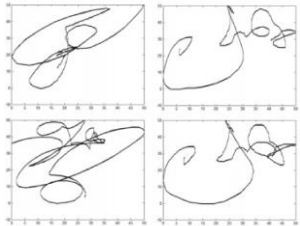

Figure 7 Examples of Genuine (first row) and Forgery (second row) signatures in ATVS dataset

ATVS has two part, named "Direct modification of the time functions" and "Modification of the sigma-log-normal parameters (LN-Parameters)". In Modification time functions, the time functions of the master signature is changed to generate the duplicated samples [4]. In modification LN-Parameters, duplicated samples are generated modifying the log-normal parameters of the master. Both methods use 25 signatures from 350 users.

2. Evaluation Parameters

Different evaluation parameters have been used in verification systems. In the following, a short description of most commonly used parameters has been summarized.

1) **Receiver Operating Characteristic (ROC) Curve**: A one-class classifier can be evaluated based on small fraction false negative (false reject rate) and false positive (false accept rate). ROC curve shows how the fraction false positive varies for varying fraction false negative.
2) **Equal Error rate (EER)**: Based on the ROC curve of a classifier, EER can be defined such that false positive and false negative fractions are equal. This parameter is a simple way to compare system accuracies.
3) **Area Under the ROC Curve (AUC)**: AUC is one way to summarize an ROC curve in a single number. This integrates the fraction true positive over varying thresholds (or equivalently, varying fraction false positive).

3. Feature Learning

In feature learning phase, a methodology has been set to learn features based on signatures except of test and train sets. Therefore, all of the signatures in ATVS dataset have been used for feature learning using auto-encoder. The size of hidden layer and iteration value of auto-encoder have been selected based on an experiment with hidden size of 500 up to 3000 nodes in which the iteration value was set from 100 to 700. Finally, based on experimental results that described in the next subsection, an auto-encoder comprises one hidden layer with 2000 nodes and the limited Broyden-Fletcher-Goldfarb-Shanno algorithm (L-BFGS) method with 700 iterations for minimization function have been chosen.

4. Classification and Verification

In this phase, SVC2004 and SUSIG datasets have been used for a K-Fold Cross-Validation process that has been implemented to categorize train and test signature groups. Several experiments have been done to achieve the best values for system parameters. EER and AUC results for SVC2004 and SUSIG datasets have been shown in Figure 8 and Figure 9, respectively.

The results shown a decrement in EER and an increment in AUC rate while facing iteration value increment. Due to change mitigation in more than 700 iterations, the iteration value has been set to 700. Although for hidden size parameter, the rate of EER enhancement and AUC rate decreased for hidden sizes larger than 2000 while computational costs increased and had been prone to over fitting and curse of dimensionality. Finally, the size of 2000 has been selected because of its computational efficiency and appropriate accuracy.

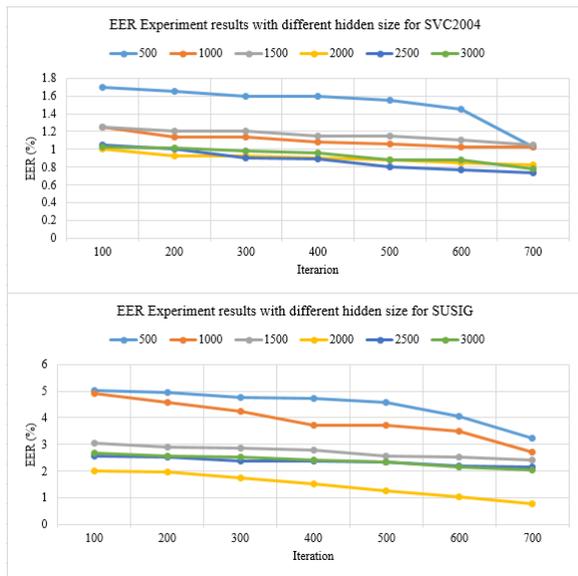

Figure 8 EER Experiment results with different hidden size for SVC2004 and SUSIG

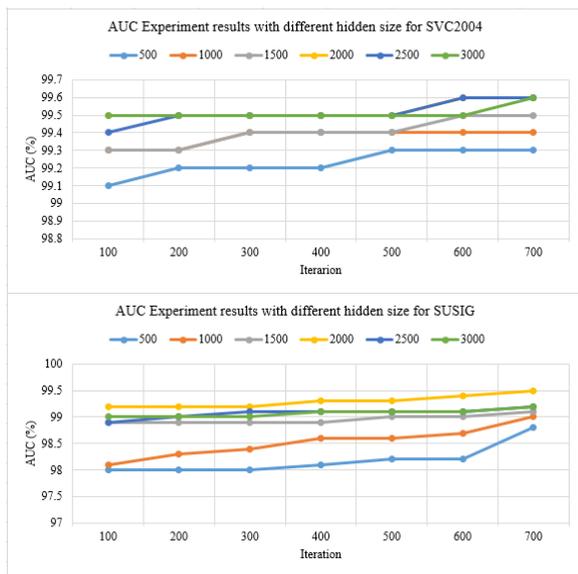

Figure 9 AUC Experiment results with different hidden size for SVC2004 and SUSIG

As a comparison between the proposed system and other approaches, verification protocols must be similar. Based on random and skilled forgery verification protocol [19, 32], 25 percent of each user's genuine signatures have been used for training to create the user model. The remaining 75 percent of user's genuine signatures, all of the skilled forgery signatures of his/her and all of the genuine signatures of other users have been used for testing based on a local threshold for each user. For evaluating the proposed method, multiple classifiers have been tested based on author's previous work [8]. These classifiers are available in Matlab open source Data Description toolbox[1] (dd_tools). This toolbox has the ability of obtaining optimal coefficients for classifiers. Finally, based on achieved experimental results, Gaussian classifier has been used.

The results of proposed method in comparison with state-of-the-art methods for two standard benchmarks (SVC2004 and SUSIG) are shown in Table 2.

Table 2. Different online signature verification methods for SVC2004 and SUSIG

| Method | SVC2004 EER (%) | SUSIG EER (%) |
|---|---|---|
| Gruber, et al. [12] | 6.84 | |
| Mohammadi and Faez [23] | 6.33 | |
| Barkoula, et al. [3] | 5.33 | |
| Yahyatabar, et al. [31] | 4.58 | |
| Khalil, et al. [18] | | 3.06 |
| Napa and Memon [24] | | 2.91 |
| Yeung, et al. [32] | 2.89 | |
| Fayyaz, et al. [8] | 2.15 | |
| Kholmatov and Yanikoglu [19] | | 2.10 |
| Ansari, et al. [2] | 1.65 | 1.23 |
| Ibrahim, et al. [13] | | 1.59 |
| **Proposed Method** | **0.83** | **0.77** |

This table indicates that proposed method have the best performance in comparison with competing algorithms. This method's EER on SVC2004 dataset is 0.83 percent, where the next best method is 1.65 percent reported for the method Ansari, et al. [2]. This verification system is 0.82 percent superior to the otherwise best result. On SUSIG benchmark, implemented method's EER is equal to 0.77 percent as it is 0.46 percent superior to the next best method.

Table 2 illustrates that in contrary to all reported methods, the results on two datasets are very close (0.06 percent difference in EER). This similarity is indicated that proposed method is dataset invariant.

## VI. CONCLUSIONS AND FUTURE WORK

In this paper, a new approach has been introduced based on self-thought learning to verify the signatures. As it can be inferred from experimental results and inherited properties of self-thought learning, the proposed system is independent from a specific benchmark, which means that it is signature shape

---

[1] Available at http://www.prtools.org

invariant.

The features, which are used to verify the signatures, have been learned from ATVS dataset by using a sparse auto-encoder with one hidden layer. By applying convolution and pooling methods, system has achieved pooled convolved features to verify the signatures. In addition, one-class classifier has been applied as it models the signatures of each user.

To compare with similar approaches, two standard benchmarks have been used which are named as SVC2004 and SUSIG datasets. Our results have shown superiority on both datasets. The features have been used in this paper can be used in other benchmarks, as this is the main component of the method proposed in this paper.

This method has proved its ability to learn the best set of features in problems that need to define handcrafted features. Therefore, it can be used in a wide range of machine learning problems. As a future work, this method can be tested on offline signatures. In addition, the impact of deep convolutional networks can be tested on both online and offline signature datasets.